\documentclass[10pt,journal,compsoc]{IEEEtran}

\ifCLASSOPTIONcompsoc
  \usepackage[nocompress]{cite}
\else
  \usepackage{cite}
\fi

\usepackage{graphicx}
\usepackage[margin=1in]{geometry}
\usepackage{cite}
\usepackage{calrsfs}
\DeclareMathAlphabet{\pazocal}{OMS}{zplm}{m}{n}
\newcommand{\La}{\pazocal{L}}
\newcommand{\Ma}{\pazocal{M}}
\newcommand{\Xa}{\pazocal{X}}
\newcommand{\Ya}{\pazocal{Y}}

\usepackage{amsmath}
\usepackage{ragged2e}

\usepackage{pifont}
\usepackage{diagbox}
\usepackage{enumitem}
\usepackage{hyperref} 
\usepackage{xcolor}
\hypersetup{
    colorlinks,
    linkcolor={red!50!black},
    citecolor={blue!50!black},
    urlcolor={magenta}
}
\usepackage[all]{hypcap}
\makeatletter
\AtBeginDocument{\let\hl\@firstofone}
\makeatother

\begin{document}

\title{Classification Auto-Encoder based Detector against Diverse Data Poisoning Attacks}

\author{Fereshteh~Razmi,
        and~Li~Xiong,~\IEEEmembership{Fellow,~IEEE}
\IEEEcompsocitemizethanks{\IEEEcompsocthanksitem F. Razmi, and L.Xiong are with Department
of Computer Science, Emory University, Atlanta,
GA, 30322.\protect\\

E-mail: {frazmim, lxiong}@emory.edu
\IEEEcompsocthanksitem L.Xiong is also with Department of Biomedical Informatics, Emory University.}
}

\markboth{IEEE Transactions on Dependable and Secure Computing, Submitted}%
{F. Razmi \MakeLowercase{\textit{et al.}}: Classification Auto-Encoder based Detector against Diverse Data Poisoning Attacks}

\IEEEtitleabstractindextext{%
\justify{\begin{abstract}
Poisoning attacks are a category of adversarial machine learning threats in which an adversary attempts to subvert the outcome of the machine learning systems by injecting crafted data into training data set, thus increasing the resulting model's test error. The adversary can tamper with the data feature space, data labels, or both, each leading to a different attack strategy with different strengths. Various detection approaches have recently emerged, each focusing on one attack strategy. The Achilles heel of many of these detection approaches is their dependence on having access to a clean, untampered data set. In this paper, we propose CAE, a Classification Auto-Encoder based detector against diverse poisoned data. CAE can detect all forms of poisoning attacks using a combination of reconstruction and classification errors without having any prior knowledge of the attack strategy. We show that an enhanced version of CAE (called CAE+) does not have to rely on a clean data set to train the defense model. 
The experimental results on three real datasets (MNIST, Fashion-MNIST and CIFAR-10) demonstrate that our defense model can be trained using contaminated data with up
to 30\% poisoned data and provides a  significantly stronger defense than existing outlier detection methods. The code is available at \url{https://github.com/Emory-AIMS/CAE}.
\end{abstract}}

}

\maketitle
\IEEEdisplaynontitleabstractindextext
\IEEEpeerreviewmaketitle
\IEEEraisesectionheading{\section{Introduction}\label{sec:introduction}}

\IEEEPARstart{P}{oisoning} attacks are attacks at {\em training time} \cite{biggio2013security} in which an attacker manipulates a small fraction of the training data in order to corrupt the model. Consequently, the model may learn a significantly different decision boundary, resulting in drastic test error. 

Poisoning attacks are acquiring increasing importance by emerging crowd-based systems that collect data from outside sources \cite{estelles2012towards,tahmasebian2020crowdsourcing,fang2021data}. In crowd-sourcing platforms, the attacker can cause massive damages without having a direct access to the system, but rather by poisoning the collected data from her. A few examples are autonomous driving cars, health systems, online review systems, and malware/spam detection systems. 
****** Also, recently poisoning attacks have been widespread in federated learning systems \cite{shejwalkar2021manipulating,shejwalkar2022back,sun2021fl}.

The most recognized poisoning attacks are \textit{label flipping} and \textit{optimal attacks} \cite{xiao2012adversarial,biggio2012poisoning}. In these types of attacks, according to the attacker's goal and his accessibility to the data, he may change the labels of some training samples or distort the feature space of the samples, usually in an optimal way to diverge the training from its regular path. Another class of poisoning attacks is backdoor attacks in which the attacker only targets a group of test data that include a specific backdoor trigger \cite{gu2019badnets,borgnia2021strong}. Backdoor attacks are not explored in this paper, but they can be considered as future work.

Several defense methods have been recently developed to address flipping or optimal poisoning attacks. Most of them consider poisoned points as outliers and utilize outlier detection techniques. 
They can be based on 
k-Nearest Neighbor (kNN) algorithms that consider a point with contrasting label with nearby samples as a poison \cite{paudice2018label}.
They can determine whether a point is poisoned by comparing its distance to a nearby point or other data points in its cluster \cite{paudice2018detection,laishram2016curie}. 
However, they have several limitations. First, they may only work for a particular type of attacks (optimal or flipping) as the detection is based on the change of either labels or features. Second, they rely on purely clean data to learn the patterns of normal points. Training on tainted data is plausible only when the fraction of the anomalous data is negligible. Also they usually rely on a threshold to determine an outlier.

A potential solution for outlier detection that has not been explored for data poisoning attacks is auto-encoder based method \cite{baldi2012autoencoders, vincent2010stacked} which learns the data representation in an unsupervised way. It has been utilized for generating poisoning attacks \cite{yang2017generative,feng2019learning,chan2020poison}, anomaly detection \cite{sakurada2014anomaly, zong2018deep}, and adversarial example detection \cite{meng2017magnet}. 
While promising, utilizing auto-encoders for detecting poisoned points under poisoning attacks present several challenges. First, existing methods train auto-encoders using clean data while there is no guarantee of purely clean data under poisoning attacks \cite{madani2018robustness,chen2021pois}. Second, existing methods typically select a threshold by allowing certain percentage of clean points to pass (e.g., 90\% clean data) but there is no access to such clean data under poisoning attacks. Finally, existing methods for detecting adversarial examples during inference time only utilize feature space (adversarial examples do not have labels). Thus, if they are leveraged in the context of poisoning attacks, they overlook some essential aspects of the attacks, i.e., the labels of the poisoned data (they may be flipped).  \\

\noindent\textbf{Contributions}.
In this paper, we develop a Classification Auto-Encoder based detector (CAE) that utilizes both feature space and label (class) information to defend against diverse poisoned data. We use a Gaussian Mixture Model for discriminating poisoned points from clean data so that it does not require any explicit threshold. We further propose an enhanced version of our method (CAE+) which does not require purely clean data for training.
We elaborate our contributions as follows: 

\begin{itemize}
\item We develop a classification auto-encoder based detector (CAE) to defend against diverse data poisoning attacks, including flipping and optimal attacks. The key idea is to utilize two components, a reconstruction part for learning the representation of the data from the feature space and a classification part for incorporating classification information into the data representation so it can better detect the poison points.

\item We further propose an enhanced model CAE+ so that it can be trained even on partially poisoned data. The key idea is to add a reconstruction auto-encoder (RAE) with CAE to form a joint auto-encoder architecture combined with early stopping of CAE so that it does not overfit the poisoned data while still learning useful representations of the clean data. 

\item We evaluate our method using three large and popular image datasets and show its resilience to poisoned data and advantage compared to existing state-of-the-art methods.  
Our defense model can be trained using contaminated data with up to 30\% poisoned data and still works significantly better than existing outlier detection methods. 
\end{itemize}

\section{Background and Related Work}

\subsection{Poisoning Attacks}
\label{sec:Poisoning Attacks Section}
\noindent Assume distribution $R$ on $\Xa\times\Ya$ where $\Ya=\{-1,1\}$. For a clean training dataset $D_{tr} = \{(x_i,y_i)\subseteq R \}_{i=0}^{n_{tr}}$, the goal of a binary classification task $\Ma$ parameterized by $\pmb{w}$ is to minimize objective loss function $\La(D_{tr},\pmb{w})$, w.r.t its parameters $\pmb{w}$.
In a poisoning attack, the attacker's goal is to produce $n_p$ poisoned data points $D_p = \{(x_i,y_i)\subseteq R\}_{i=0}^p$ so that using new training data $D'_{tr} = D_{tr} \cup D_p$ by the learner results in attacker's goal or objective function. This goal can be maximizing the loss on the entire clean test dataset (untargeted attacks) or on a subset or class of them (targeted attacks).

Poisoning attacks have different manifestations depending on which part of the data is manipulated during the attack. Each of them can have a different impact on attacker's objective function and different attack strength. 

In \textit{Label flipping attacks} or in short flipping attacks, only class labels of poisoned data are flipped, and the adversary usually has a limited budget for the number of samples it is allowed to change their labels \cite{zhao2017efficient,xiao2012adversarial,xiao2015support,paudice2018label}. 

\textit{Optimal attacks} are based on optimizing the poisons to drastically degrade the model's performance. These attacks are stronger compared to other poisoning attacks, since both feature space and labels can be changed. 

For classification problems \cite{biggio2012poisoning,xiao2015feature,munoz2017towards}, the rule of thumb is to initialize poisons with real samples from training data set and flip their labels. Since labels are not differentiable, they only optimize the feature space. In this paper, we also introduce \textit{Semi-optimal attacks} which keep the original labels of the points without flipping them and only optimize the feature space. This attack can be realistic when the attacker has no control over the labeling process.

\subsection{Defense against Poisoning Attacks}
\noindent\textbf{Outlier Detectors}.
Outlier detection methods are common to defend poisoning attacks. These methods are based on the fact that poisoned data deviates from normal points or underlying data generation mechanism. \cite{paudice2018detection} suggests distanced-based outlier detection methods to mitigate the effect of optimal attacks, assuming they have access to a trusted dataset to train the outlier detector. \cite{steinhardt2017certified} uses a centroid-based outlier detector and estimates a data-dependent upper-bound on the objective loss for poisoning attacks, offering a certified defense. \cite{chen2021pois} benefits from combining Generative Adversarial Network (GAN) based models, namely cGAN and WGAN-GP, to create augmented clean samples from a trusted dataset and mimic the original model. It then compares the training data against a threshold calculated by the augmented clean data to detect the poisoned samples. 
Outliers can also be detected by clustering based methods \cite{shen2016auror}. \cite{laishram2016curie} considers clustering in the combination of feature and label space to defend against optimal attacks and showed that the poisoned points are more separated from the rest of their cluster compared to using features alone.
K-nearest neighbor algorithm is proposed to combat flipping attacks \cite{paudice2018label}. They assume samples close together share a common label; otherwise, the sample's label is highly likely flipped. 

As we discussed, the existing outlier detection based methods have several limitations.  They typically focus on one type of attacks and rely on purely clean data to train the detector and a threshold to determine outliers and hence are not very effective or robust.
Furthermore, a recent work \cite{koh2021stronger} considers a new attack method that generates adjacent poisoned samples. In this case, proximity-based outlier detection algorithms such as K-nearest neighbor fail to recognize poisoned data.\\

\noindent\textbf{Contribution-based Methods}. Another type of defense methods are based on how and to what extent each point contributes to or influences the resulting model. In the context of regression problems, \cite{jagielski2018manipulating} retrains the model multiple times and removes points with high residuals as poisonous points.
RONI \cite{nelson2008exploiting} tests the impact of each data point on training performance and discards those points that have a negative contribution. \cite{baracaldo2017mitigating} takes a similar approach but reduces computational expenses by examining the impact of entire group of points on the model in order to find manipulated groups. \cite{koh2017understanding} is another work that attempts to find the high impact samples in a less costly process using influence functions. \cite{hong2020effectiveness} shows poisoned data influence the magnitude and orientation of gradients during the training. They use DP-SGD \cite{abadi2016deep}, a method of differential privacy \cite{zhu2017differential}, to add noise to the gradients and limit their sensitivity to mitigate the impact of poisoned samples.
In summary, while showing promising results without approximation, the contribution-based methods have the main drawback of high computation cost due to enumerative retraining, which makes it impractical, especially for settings where potentially poisonous data are being continuously acquired. Also, for the differentially private defense methods, they suffer from a decrease in model utility \cite{hong2020effectiveness}. 

\subsection{Auto-encoders in Anomaly Detection}
\noindent Auto-encoders
\cite{baldi2012autoencoders, vincent2010stacked} are neural networks that learn data representation in an unsupervised way.   
 Auto-encoders reconstruct input $x$ into output $x'$ by minimizing reconstruction error usually on an $L_p$-norm distance:
\begin{equation}
  RE(x) =  \parallel x - x'\parallel_{p}
  \label{eq:re}
\end{equation}
If auto-encoders are trained with only benign data, they learn to capture only the main characteristics of these data. So when the reconstruction error of a sample exceeds a threshold, it is considered an anomaly \cite{an2015variational, sakurada2014anomaly}. Nevertheless most anomalies are recognized as samples with observable differences from the real data \cite{aytekin2018clustering}, which are not effective for poisoned data that have very small perturbations. 

 Auto-encoders have been proposed to detect adversarial examples at inference time by Magnet \cite{meng2017magnet}.  
In addition to considering reconstruction error between the input and output, they also feed them to the target classifier and compare the corresponding softmax layer outputs to boost the detection power. However, in the context of poisoning attacks, a pre-trained trusted classifier does not exist. Instead the defender has access to an extra piece of information which is the associated label of the poisoned point. 

For both outlier detection and adversarial example detection, the auto-encoders need to be trained with pure clean data to capture shared properties amongst normal data 
\cite{meng2017magnet, an2015variational}. Even in some works \cite{zhou2017anomaly, aytekin2018clustering} that considered anomalous data in the training process of the auto-encoders, the percentage of anomalies in the dataset is insignificant. In the setting of poisoning attacks, the assumption of having a clean dataset for training the defense method is not realistic. For training our defense model, we assume 10\% of poisoned points. By utilizing a joint architecture, we show that our defensive model can remain resilient to poisoning attacks. 

\section{Classification Auto-encoder Based Detector}
\label{section:poisonedDatection}

As a baseline solution, we can train an auto-encoder on feature space as in existing outlier detection methods. For a clean sample $s_c=(x_c,y_c)$, and a poisoned sample $s_p=(x_p,y_p)$, $RE(x_p)$ can be used to discriminate $x_c$ and $x_p$:
\begin{equation}
  RE(x_c) << RE(x_p)
  \label{eq:re_re}
\end{equation}
According to (\ref{eq:re_re}) any data point with significantly large reconstruction error can be considered as a poison. The limitation of this approach is that it will only capture the changes in the feature space.
Hence it will address only semi-optimal attacks which only change the features.
\begin{figure*}
\centering
\setlength{\belowcaptionskip}{-10pt}
\includegraphics[width=1\linewidth]{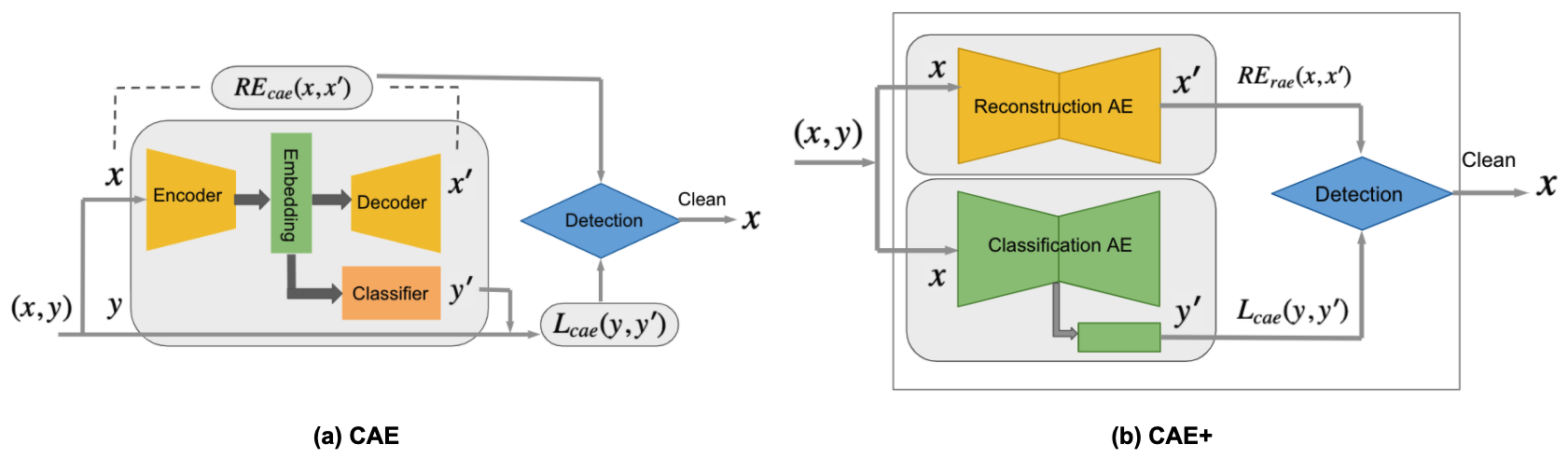}
\caption{Auto-encoders Structure: (a) The structure of Classification Auto-encoder (CAE). If trained on pure clean dataset it provides a high success defense against all poisoning attacks. (b) The structure of CAE+. Both Reconstruction Auto-encoder (RAE) and Classification Auto-encoder (CAE) work together to combat against poisons. This joint structure makes the defense method more robust even if trained on a contaminated dataset.
}
\label{fig:modelSketch}
\end{figure*}

\subsection{Classification Auto-encoder (CAE)}

\noindent To defend against all types of poisoning attacks, we need a method that incorporates both labels and features in detection process. In other words, the latent encoding of the auto-encoder needs to reflect the label information.  \\

\noindent\textbf{Classification Auto-Encoder.}
We propose Classification Auto-Encoder (CAE) which has an auxiliary classifier fed by the latent representation $z$ of the encoder (Figure~\ref{fig:modelSketch}(a)).
If $RE_{cae}$ indicates the reconstruction error,
and $L_{cae}$ indicates the auxiliary classifier's loss on representation layer $z$, while training the CAE, it tries to minimize $\sum_{x_{i}} (RE_{cae}(x_i)+L_{cae}(x_i)))$ on training dataset $D_c=\{x_i,y_i\}_{i=0}^n$. 
As a result, $z$ is learned in such a way that the classifier is able to predict the label, and the  decoder can reconstruct the associated input. To boost the connection between these two tasks, we train the auxiliary classifier and the decoder simultaneously. It contrasts with previous works that utilize classification auto-encoders for predictive or classification objectives. They employ a two-stage training process; first, they train the pair of encoder-decoder and then use the low-dimensional representation for training the classifier \cite{xing2016stacked, geng2015high}.\\

\smallskip
\noindent\textbf{Detection Criteria.}
Once the CAE is trained, given a data point, we can use the combined reconstruction error and classification loss as a detection criteria for poisoned data, since it considers deviations in both feature space and label space. 
\begin{equation}
  Error
  (x) = \alpha . RE_{cae}(x) + (1-\alpha) . L_{cae}(x)
  \label{eq:error_cae_clean}
\end{equation}

The first term $RE_{cae}(x)$ is the reconstruction error of CAE and the second term $L_{cae}(x)$ is the loss of the CAE auxiliary classifier. $\alpha$ and $1-\alpha$ are weights to control the effect of each term. 
Since $RE(x)$ is indicative of changes in $x$, and $L(x)$ reflects the classification loss, the combined metric $Error(x)$ can detect both changes in feature space and labels and hence defend against the different types of attacks.

\label{section:GMM}

In general, a threshold can be defined based on \textit{a guess on the number of possible poisoned points} $K$ \cite{jagielski2018manipulating}.
Tuning $K$ is a difficult job that makes the detector very sensitive to the actual fraction of poisoned data. Instead, we use a clustering approach and cluster all points based on $Error(x)$ into two components using a Gaussian Mixture Model (GMM). We show that the error is so distinct between clean and poisoned points that GMM can separate it very well into two clusters, each representing clean or poisoned data.

\subsection{Enhanced Classification Auto-encoder (CAE+)}

\noindent CAE requires clean data for training the auto-encoder so it can learn the structure of the normal data and detect any deviation from that. Since we assume the training data is poisoned, we need to add a mechanism that is robust to contaminated data. We do so by leveraging a combination of early stopping method and a replicate reconstruction auto-encoder.\\

\noindent\textbf{Early Stopping}.
Since we assume there is no access to purely clean data for training the detector, to prevent CAE to learn patterns from poisoned data, we use the early stopping method. Early stopping leads the auto-encoder to focus on reconstructing the pattern of the majority of data, and avoids overfitting on anomalies.
The auxiliary classifier is a single dense layer and can usually catch all the class information quickly, especially in binary-class problems. Selecting a small number of neurons in this layer does not provide sufficient parameters for the classification task, and leads to missing even the general patterns of the training dataset. On the other hand, large number of neurons makes the classifier more complex and may overfit the poisonous data. To capture all the information and avoid underfitting, we can select a fairly large number of neurons and address the overfitting problem using early stopping.

By using this approach, CAE can be very robust to the poisoned data. However, at the stop point of the training process, $z$ has captured those patterns of the data that help mostly with classification, but not the reconstruction  (which takes longer to learn). 
Hence we propose a joint auto-encoder architecture to address this challenge by using a parallel reconstruction auto-encoder (RAE).\\

\noindent\textbf{Reconstruction Auto-encoder}.
The Reconstruction Auto-Encoder (RAE) is a replicate of the encoder-decoder part of CAE without the classification layer. RAE is trained to minimize the reconstruction error only. 
By having these two auto-encoders, for an input $\{x,y\}$ we calculate the following combined error:
\begin{equation}
  Error(x) = \alpha . RE_{rae}(x) + (1-\alpha) . L_{cae}(x)
  \label{eq:error_cae}
\end{equation}

This is a modification to (\ref{eq:error_cae_clean}), in which the reconstruction error has been replaced with reconstruction error of RAE ($RE_{rae}(x)$).

This extra auto-encoder helps us adjust the training process for RAE separately so that while RAE can be trained to full capacity, CAE is not overfitting the poisonous data using early stopping. In comparison to the classifier of CAE, RAE with high capacity (especially with convolutional layers) can be trained with a high number of epochs without overfitting the poisoned data.
We call this joint structure of CAE and RAE, CAE+, since it is enhancing the CAE functionality (Figure~\ref{fig:modelSketch}(b)).

In practice, the training data may be poisoned, so using CAE+ and Equation~\ref{eq:error_cae} is required. In Section~\ref{experiments}, we investigate potential scenario of having a clean training dataset $D_c$ and compare CAE vs. CAE+. In the case of clean training data, since the concern of overfitting on poisoned data does not exist, CAE can be trained until both the classification layer and decoder converge. We show that CAE can be effective under this circumstance. In contrast, when training data is poisoned, we show that CAE+ is much more robust. 

\section{Experiments}
\label{experiments}
In Section~\ref{sec:Experimental Setup}, we describe the details of our experimental settings, including the datasets, the attacker's target model, the architecture of our detectors, the comparison methods, and the attributes of the attacks. We also offer a fourth type of attack that combines all other poisoning attacks to show the strength of CAE+ against all kinds of attacks. Furthermore, we clarify how we used the periodic update of the model to mimic real scenarios wherein poisoning attacks occur. 

In the Results section~\ref{sec:Results Section}, we depict the impact of each type of attack on the poisoned data, the prominence of the Gaussian Mixture Model (GMM) over threshold selection, and the effect of the different auto-encoders employed in the CAE+. Then an ablation study reveals the dominance of CAE+ over CAE and RAE. To confirm the superiority of CAE+, it is compared to the other state-of-the-art detectors in the literature on multiple datasets, including CIFAR-10. Finally, in the last subsection, we illustrate the robustness of CAE+ vs. CAE when, unlike the other experiments, we assume there is a trusted training dataset for training CAE.

\subsection{Experimental Setup}
\label{sec:Experimental Setup}
\noindent\textbf{Datasets}. First, we evaluate the performance of CAE+ using the MNIST dataset \cite{lecun1999object}, and more challenging Fashion-MNIST dataset \cite{xiao2017fashion} on binary sub-problem classes: MNIST 9 vs. 8 and 4 vs. 0, and Fashion-MNIST Sandal vs. Sneaker and Top vs. Trouser. It is common practice to apply binary setting for data poisoning attacks \cite{koh2017understanding,biggio2012poisoning}. Second, we conduct experiments on a more complex dataset CIFAR-10 \cite{krizhevsky2009learning} for two randomly chosen classes Airplane vs. Automobile. All datasets are normalized within the interval [0, 1].\\

\noindent\textbf{Attacks}.
Support Vector Machines (SVM) are known to be subject to strong poisoning attacks \cite{biggio2012poisoning, xiao2012adversarial}. In contrast to complicated models and neural networks \cite{munoz2017towards}, poisoning attacks can achieve a high success in dropping the  accuracy of SVM. As we will show in Figure~\ref{fig:accplots}, the accuracy of optimal attacks on the SVM model  drops to 60\% with 10\% of poisons. Hence, we use poisoning attacks against SVMs in the experiments to better demonstrate and evaluate the effectiveness of different defense methods.  We use linear kernel for MNIST and Fashion-MNIST and RBF kernel for CIFAR-10. We note that our methods work on poisoning attacks against any target models such as neural networks.

We compare four types of attacks; flipping, optimal, semi-optimal, and mixed attacks, then assess our defense model against them. In a mixed attack, the attacker selects 1/3 of the poisons from each of the aforementioned attack types. This way, we can challenge the defender's ability to detect diverse poison simultaneously, despite their different characteristics. The optimal attack is conducted based on \cite{melis2019secml} with some modifications.\\

\begin{figure*}
  \centering
    \setlength{\belowcaptionskip}{-20pt}
    \includegraphics[width=\linewidth]{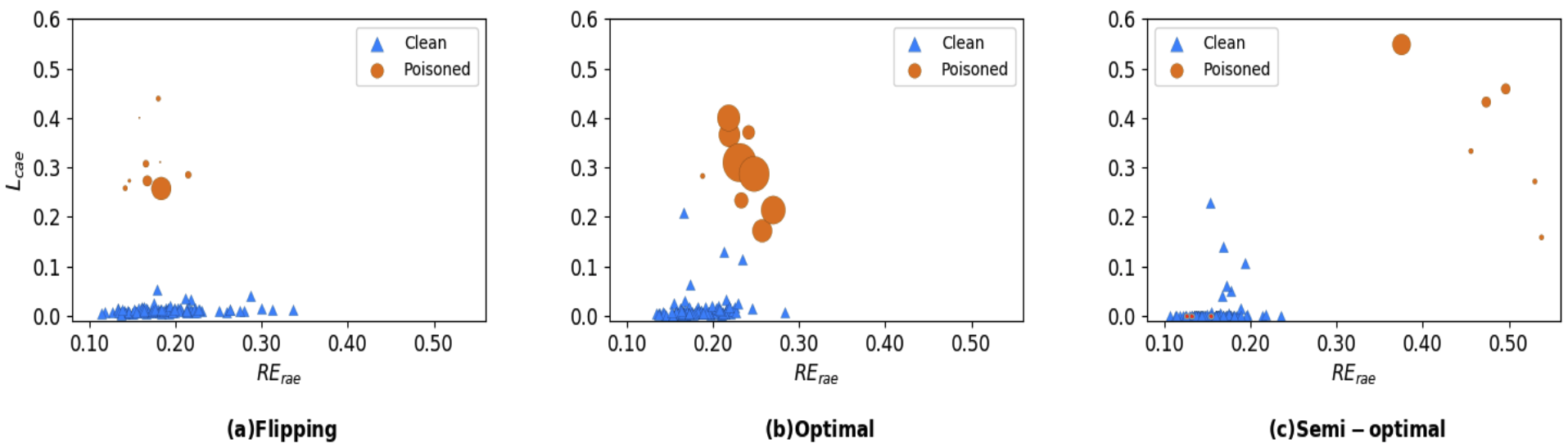}
    \caption{The effect of different attack types on the reconstruction error and auxiliary classification loss for poisoned MNIST-4-0 dataset. Triangles and circles represent clean and poisoned points, respectively. The poisons' size represents their impact on degrading the SVM accuracy (larger circles indicate higher impact).}
    \label{fig:F1_RE_Laux}
\end{figure*}

\noindent\textbf{Setup}.
A common paradigm for training ML models in real world is the periodic update \cite{laishram2016curie} in which the data is acquired continuously. In this scenario, data is provided by users and buffered until sufficient data is obtained to retrain the model. To implement such a periodic update setting for SVM classifier, we consider 60 rounds of SVM updates. Each round represents a new batch of data which consists of 500 data points divided into a training set, a validation set, and a test set of 100, 200, and 200 samples, respectively. Based on different attack types, the attacker generates poisoned points for each round and adds them into the training data  for that round. At the next step, we assume that the defender has access to the recent 50 rounds of buffered data. By aggregating the contaminated buffered data of those 50 runs, we train our defense model. Then for evaluation purposes, we use the remaining 10 rounds of updates for testing the defense methods, namely 10 times the buffered data is fed to the detector and the data passing through it is used for model assessment. Every result reported in this paper is the average of these 10 test runs.

Note that for each of the attacks unless otherwise specified, up to 10\% of the clean data are poisoned. Exceeding 10\% may not be realistic in practice \cite{koh2021stronger, carnerero2020regularisation, biggio2012poisoning, laishram2016curie}. We believe this is high enough to validate the robustness of CAE+ against poisonous data. To further show the impact of the percentage of the poisoned data, we conduct the experiment on CIFAR-10 with a higher poisoning rate (up to 30\%). \\

\noindent\textbf{Implementation Details}.
\label{Implementation}
The structure of CAE reconstruction component and RAE is inspired by the auto-encoders introduced in Magnet \cite{meng2017magnet} with some modifications. Our reconstruction auto-encoders, for MNIST and Fashion-MNIST dataset, consist of 3x3 convolutional layers in the encoder, each composed of 3 filters of size 3x3 with 1x1 strides and sigmoid activations. Between these two convolutional layers a MaxPooling 2x2 is located. At the decoder, the structure of Convolutional layers are the same as the encoder. The only difference is that the MaxPooling layer is replaced with a 2D UpSampling layer. As the last layer of the decoder we have a third 3x3 convolutional layer with only one filter (compatible to number of channels in MNIST and Fashion-MNIST) to reconstruct an output image with the same size as the input image. Also, as \cite{meng2017magnet} suggests, we use a slightly different architecture for CIFAR-10, by utilizing only one convolutional layer in the encoder and one in the decoder with the mentioned parameters.
For the auxiliary classifier, encoder's output is flattened and fed to a dense classification layer with size 128. We experimentally found out that dropping out the data with rates 0.25 and 0.5 before and after the dense layer serves the best in training the data and reduces the overfitting. For each dataset, we train CAE for 100 epochs and the RAE for 300 epochs with a batch size of 256 using the Adam optimizer. The aggregated error $Error(x)$ is calculated based on \eqref{eq:error_cae} on weighted sum of the normalized $L_1$-norm reconstruction error and the auxiliary classifier's cross entropy loss.\\

\begin{figure*}
  \centering
  \setlength{\belowcaptionskip}{-20pt}
    \includegraphics[width=\linewidth]{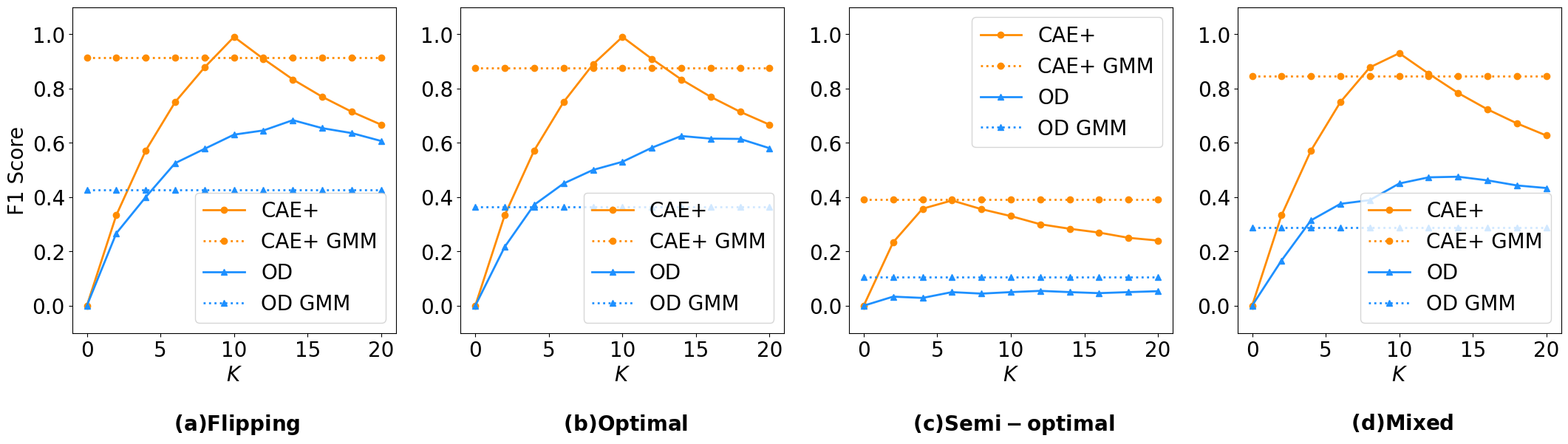}
    \caption{Changes on MNIST-4-0 F1-score over different thresholds for CAE+ and OD. Thresholds are guesses on the probable number of poisoned data within the training dataset. 
    }
    \label{fig:F1_thr}
\end{figure*} 

\begin{figure}
    \centering
    \includegraphics[width=1.0\linewidth]{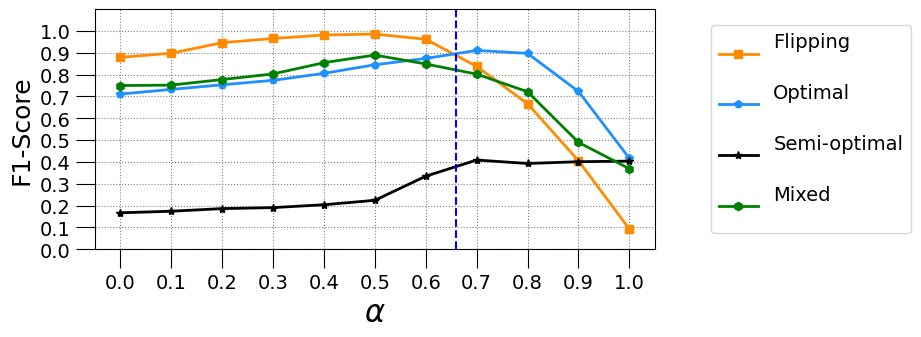}
    \caption{CAE+ F1-score for different values of $\alpha$ (Equation~\ref{eq:error_cae}). 
    }
    \label{fig:alpha}
\end{figure} 

\noindent\textbf{Comparison Methods}.
Distance-based outlier detectors are state-of-the-art methods in defending against poisoning attacks \cite{koh2021stronger,paudice2018detection}. One of their interesting properties is that they are very robust against poisoned data and do not require to be trained on a purely clean dataset. So, similar to \cite{paudice2018detection}, we select \textbf{centroid-based Outlier Detectors (OD)} as the baseline. It first finds the centroids of each class in the training dataset and then discards the points that are distant from their respective class centroid.

Furthermore, we compare our method to a modified Magnet, a state-of-the-art auto-encoder based detector designed for adversarial examples. We make the following modifications in order to make it compatible with poisoning attacks under our setting. We train Magnet on the same poisonous data as the other defense methods. It contrasts with the Magnet original paper in which the authors train the Magnet on a thoroughly clean dataset. It is based on the assumption of access to such a clean dataset, which is valid under evasion attack (against adversarial examples) at inference time, but not the poisoning attacks during training time.
We use the same structure as the original paper suggests \cite{meng2017magnet}, the only hyperparameter we change is the number of epochs for a better adaptation to poisoning attacks (from 100 epochs to 300 epochs). In addition to the detector, we also evaluate the performance of Magnet detector paired with a reformer \cite{meng2017magnet}. In this case, after Magnet detector filters out poisons, it passes the remaining data through the reformer, which is another auto-encoder. The reformer's reconstructed output will replace the original input and then be fed to the classifier.  

\subsection{Results} 
\label{sec:Results Section}
\noindent\textbf{Effect of Different Attacks}.
As we discussed in Section~\ref{section:poisonedDatection}, each type of poisoned data can have a different impact on CAE+ components. Figure~\ref{fig:F1_RE_Laux} illustrates this fact by showing the classification error $L_{cae}$ and reconstruction error $RE_{rae}$ of the different poisoning attacks on MNIST-4-0. Blue triangles and orange circles represent the clean and poisoned points, respectively. Clean data is the same for all four plots. For the poisoned data, the size of circles indicates their importance in degrading the SVM classification results. Larger circles imply that the insertion of those poisons to the SVM clean training dataset drops more accuracy. 

For the flipping attack, the reconstruction error $RE_{rae}$ cannot differentiate the poisoned samples from the rest of the data since the feature space of the poisons is intact, while the classification loss $L_{cae}$ is much larger for the poisoned data. Under the optimal and semi-optimal attacks, the transformations that occur in the feature space discriminate the clean data and the poisons through $RE_{rae}$. It is more noticeable for the semi-optimal attack because the features alter more drastically than in the optimal attack. This discrepancy between the poisons' features and the clean space impacts their classification results and increases the loss $L_{cae}$. Therefore, as Equation~\ref{eq:error_cae} suggests, a mixture of both reconstruction and classification errors is required to detect diverse attacks in the context of an attack-agnostic defense.\\

\begin{figure*}
  \centering
  \setlength{\belowcaptionskip}{-20pt}
  \includegraphics[width=\textwidth]{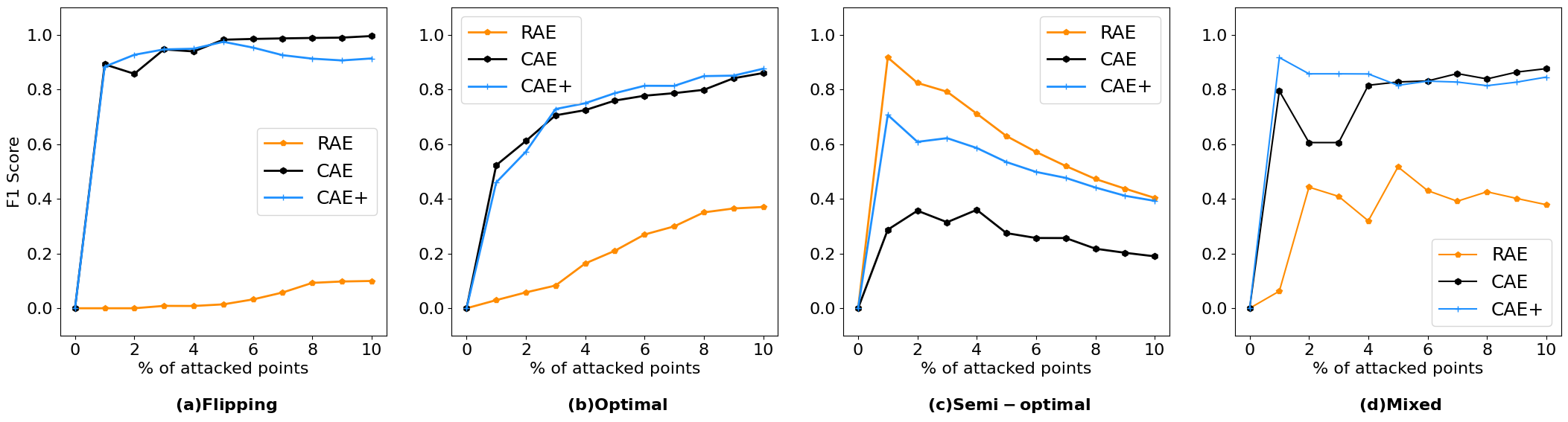}
  \caption{ Ablation study between CAE+, CAE and RAE on MNIST 4-0
  }
 \label{fig:ablation}
\end{figure*}

\noindent\textbf{Threshold vs. GMM}.
According to Section~\ref{section:GMM}, we pass the detectors' output to a GMM for clustering the data into poisoned and clean data, so that we do not need to specify a threshold of possible poisoned points $K$ for filtering poisons. We compare our GMM-based approach with the baseline threshold approach when a fixed number of training data is poisoned (about 10\% of the training data, i.e., 10 poisons).  We report \textbf{F1-score} for the detection, which is the harmonic mean of the precision and recall with the best value at 1. F1-score is indicative of how successful a detector is in filtering poisons and passing clean data. An ideal detection algorithm can identify all and only poison data, which means a perfect F1-score.

Figure~\ref{fig:F1_thr} depicts how the detectors'  F1-scores change with different threshold of $K$ for MNIST-4-0 (solid lines). For flipping, optimal and mixed attacks, the F1-score of CAE+ hits almost 1 at $K=10$. In other words, it can accurately detect all ten poisoned points with very few false positives. The V shape of CAE+ plots depicts its sensitivity to an accurate threshold $K$. Before threshold 10 there are naturally some false negatives, and after that point, false positives are emerging. In contrast, we do not need to specify any threshold in the unsupervised GMM method (dashed line) for CAE+. We can see that it competes very closely with the best guess on $K$ in the threshold-based method. 

For the semi-optimal attacks, the scenario is slightly different. The majority of the poisoned points in semi-optimal attacks get stuck in local maxima and do not change their feature space; hence they have little impact on the attack. For the same reason, they do not harm the accuracy even though they can not be filtered out. This fact is illustrated in Figure~\ref{fig:F1_RE_Laux}. Some of the low-impact attacked points (shown with small circles) are placed at the bottom left corner of the plot, where the majority of the clean data points are located. As a result, in Figure~\ref{fig:F1_thr}, F1-score for semi-optimal attacks is not high; but we show later that CAE+ can detect all the high impact attack points and achieve the original SVM's accuracy. 

In all the attacks,  for both threshold-based and GMM methods, CAE+ yields significantly better F1-scores than OD. For linear SVM, overlooking poisoned points can be much more harmful than filtering out clean data. So despite the high false-positive rate, OD can still partially enhance the SVM accuracy. OD completely fails to operate as a detector if the system is sensitive to clean data removal. In the remaining experiments, we leverage GMM for all the detection approaches to have a fair comparison of how they boost SVM accuracy.\\

\begin{figure*}
  \centering
  \setlength{\belowcaptionskip}{-10pt}
    \includegraphics[width=\textwidth]{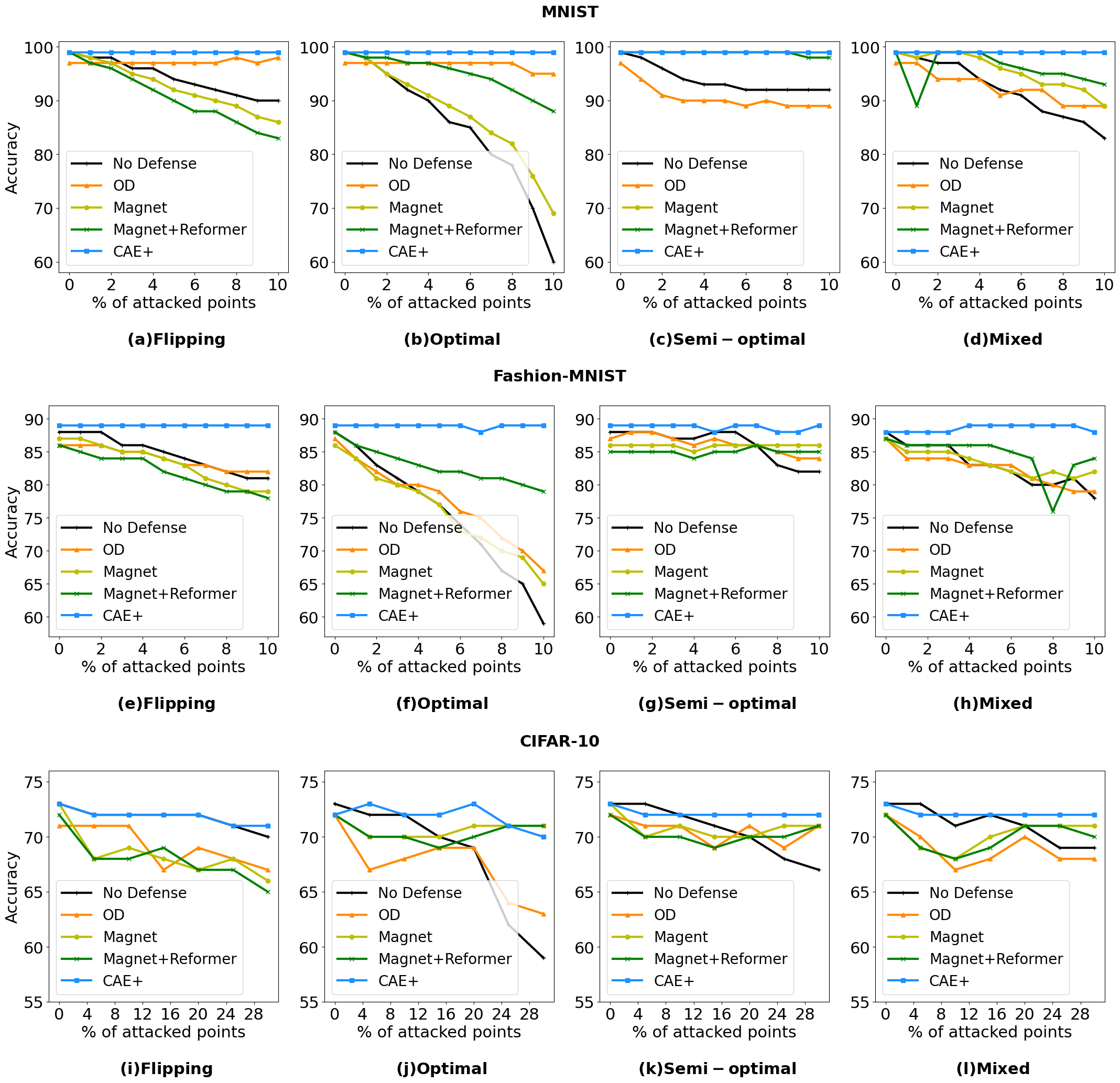}
  \caption{Comparison of SVM accuracy after filtering suspicious points by CAE+, OD, and Magnet over different percentages of poisons. The first row represents MNIST-4-0, the second row is Fashion-MNIST Sandal-Sneaker and the third row belongs to CIFAR-10 Airplane-Automobile. 
  }
 \label{fig:accplots}
\end{figure*}

\begin{figure*}
  \centering
    \setlength{\belowcaptionskip}{-30pt}
    \includegraphics[width=\linewidth]{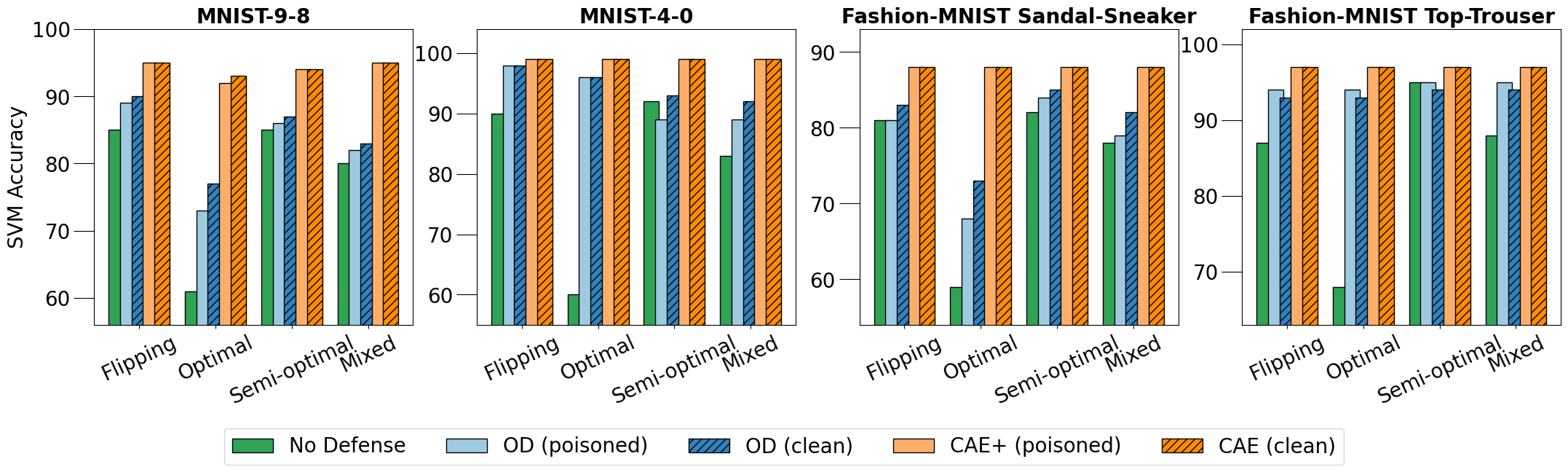}
    \caption{Comparison of SVM accuracy using detectors trained on clean vs. poisoned data. 
    }
    \label{fig:clean-pois-acc}
\end{figure*}

\noindent\textbf{Impact of Alpha}.
There are four types of attacks. Each of the CAE+ reconstruction or classification auto-encoders is suitable to address different attack types. Coefficient $\alpha$ in Equation~\ref{eq:error_cae} can be adjusted to meet this goal. Since the attacker's attack type is not known to the defender, $\alpha$ should be pre-adjusted considering all the attack types. Figure~\ref{fig:alpha} demonstrates how different values of $\alpha$ affect F1-score. Reconstruction error has a significant impact on semi-optimal attacks, and as a result, higher $\alpha$ boosts the F1-score. In flipping attacks and optimal attacks, classification error gains more importance. In particular, in optimal attacks, there is a trade-off between reconstruction error and classification error. 
The vertical dashed line shows $\alpha$=0.66 in which every attack sustains high F1-socre. According to Equation~\ref{eq:error_cae}, at this value the coefficient of $RE(x)$ is twice as the coefficient of $L(x)$.\\

\noindent\textbf{Ablation Study}.
In this section we show the contribution of each component in CAE+ (recall Figure~\ref{fig:modelSketch}). 
We train two additional models for comparison: 1) CAE that is not combined with the RAE (the bottom auto-encoder in Figure~\ref{fig:modelSketch}(b)) and has the error function in Equation~\ref{eq:error_cae_clean};  2) RAE that is a stand-alone reconstruction auto-encoder (the top auto-encoder in Figure~\ref{fig:modelSketch}(b)) and uses reconstruction error as defined in Equation~\ref{eq:re}.

The error for CAE is calculated based on Equation~\ref{eq:error_cae_clean}, and for RAE, it is limited to just reconstruction error. Note that all these methods are trained with 10\% contaminated data and paired with GMM. 
Figure~\ref{fig:ablation} shows the effectiveness of these detectors based on F1-score. Since RAE considers only feature space, it is effective on semi-optimal attacks and, to a less extent, on optimal attacks. However, flipping attacks can evade it. On the other hand, CAE relies on classification and reconstruction errors with more emphasis on classification loss. So it fails on semi-optimal attacks. CAE+ has the advantage of using both CAE classification error and RAE reconstruction error, and as a result, it gains a better F1-score on average. Since the attack is not known in advance, CAE+ is the best detector among these three.\\

\noindent\textbf{Comparison}.
In this experiment, we compare the performance of CAE+ in terms of accuracy of the resulting model with state-of-the-art defense methods.
We feed the learner's training data into detectors and filter suspicious poisoned points using GMM. The rest of the points are used to retrain the SVM classifier. A perfect filter leaves us with the entire clean data, excluding all poisons, which results in a high SVM accuracy.\\

Figure~\ref{fig:accplots} illustrates the resulting accuracy on different percentages of poisoned training datasets. The plots on the first row (a to d), second row (e to h) and third row (i to l) belong to MNSIT-4-0, Fashion-MNIST-Sandal-Snkear and CIFAR-10 Airplane-Automobile, respectively, with original accuracies of 99\%, 88\% and 73\% on clean unpoisoned datasets. In each row, all plots have the same scale. Each plot indicates one type of attack and corresponding detection methods.

In each plot, we show the accuracy without any detection (attack), and the accuracy with CAE+, in comparison with other three detection methods (OD, Magnet, and Magnet+reformer).
We first elaborate on the results of the first row, for MNIST-4-0 dataset. Considering each plot individually, for all the attack types, CAE+ constantly achieves almost the original accuracy (blue lines), and outperforms other detectors. As expected, optimal attacks are the strongest among all four types of attacks.

Magnet does not consider label flipping, so it fails on flipping attack scenarios. When the feature space changes are significant (mostly semi-optimal attacks), its performance is comparable to CAE+. Magnet's sensitivity to perturbation size has been explored in \cite{meng2017magnet} for evasion attacks under multiple adversarial example distortion rate $\epsilon$. Adding the reformer enhances Magnet's results significantly. It gives us the insight that using the reformer along with CAE+ can boost its performance. We did not experiment on CAE+reformer, but it can be a direction for future work. The Fashion-MNIST and CIFAR-10 results are similar to MNIST.

Note that MNIST and Fashion-MNIST were tested for up to 10\% of poisoned data, and CIFAR-10 is tested for up to 30\% of poisons. 
Although it is not practical for an attacker to inject this high number of poisons into the system in real world, but this is a good stress test to show CAE+ is robust to even higher poison rates.\\

\noindent\textbf{Robustness}.
Given the assumption of having access to only an untrusted (contaminated) dataset, CAE+ was chosen over CAE in all the previous experiments. However, if clean data is available, we can simply use CAE. Therefore, to verify the impact of this assumption on the detectors' performance, we train a stand-alone CAE on clean data, utilizing (\ref{eq:error_cae_clean}) and on a large number of epochs (300). In this experiment, we end up with two new detectors; a clean CAE and a clean OD. 

We use a training dataset with 10\% poisoned data to train SVM, then apply both clean and poisoned versions of CAE(+) and OD on this data to see how they filter poisoned points and recover SVM accuracy. The result of this comparison on four datasets MNIST 9-8, 4-0, Fashion-MNIST Sandal-Snkear, and Top-Trousers are represented in Figure~\ref{fig:clean-pois-acc}. The original SVM accuracies on trusted data for these datasets are 95\%, 99\%, 88\%, and 97\%, respectively. We observe that OD is susceptible to contaminated data as clean OD usually surpasses its contaminated version. We note that when the defender has access to a clean dataset, it is adequate to train CAE directly without CAE+. Also, CAE+/CAE always outperforms OD, especially in optimal attacks. 

\section{Conclusion}
This paper utilized auto-encoders to defend against various types of poisoning attacks for the first time. We proposed CAE, a two-component auto-encoder that enjoys an auxiliary classification layer to boost detection performance. We enhanced the structure of CAE by introducing CAE+. The enhanced version is a joint auto-encoder detector that has a high robustness against contaminated data. Experiments demonstrated the detection power of CAE+ against diverse poisoning attacks including optimal, semi-optimal and label-flipping attacks and showed that it surpasses the state-of-the-art distance-based outlier detector and Magnet detector. In all these cases, CAE+ is trained on a dataset that is corrupted with a high rate of poisoned data and still preserved its performance. 

Directions for future work include demonstrating the results of such detectors on non-convex target models. Studies can also explore the influence of these defensive approaches on multi-classification problems. Besides, it is worth extending current work on backdoor attacks since they are carried out during training and share many characteristics with poisoning attacks.

\ifCLASSOPTIONcaptionsoff
  \newpage
\fi

\begingroup
\renewcommand{\section}[2]{}%
\subsection*{References}
\bibliographystyle{plain}
\bibliography{bib_file}
\endgroup

\newpage
\begin{IEEEbiography}[{\includegraphics[width=1in,height=1.25in,clip,keepaspectratio]{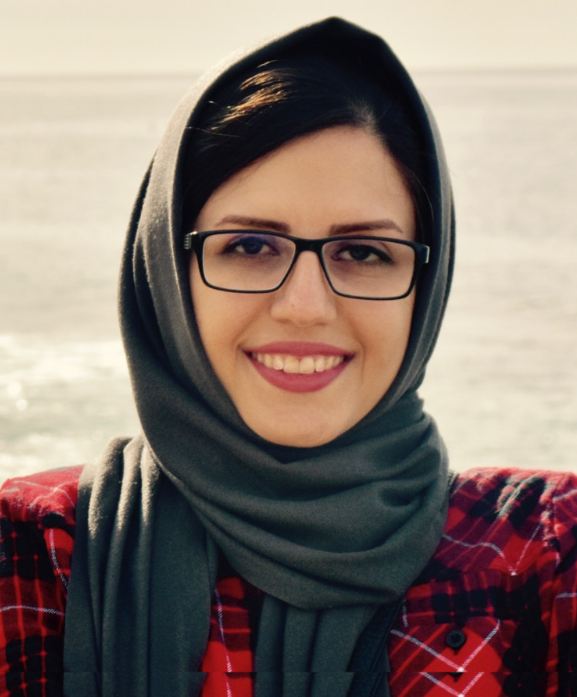}}]{Fereshteh Razmi}
Fereshteh Razmi is a fifth year Ph.D. student of Computer Science and Informatics at Emory University, mainly focusing on Adversarial Machine Learning, Poisoning Attacks and Security in Deep Learning. She is currently exploring the connection between Poisoning Attacks and Differential Privacy in deep learning models.
\end{IEEEbiography}
\begin{IEEEbiography}[{\includegraphics[width=1in,height=1.25in,clip,keepaspectratio]{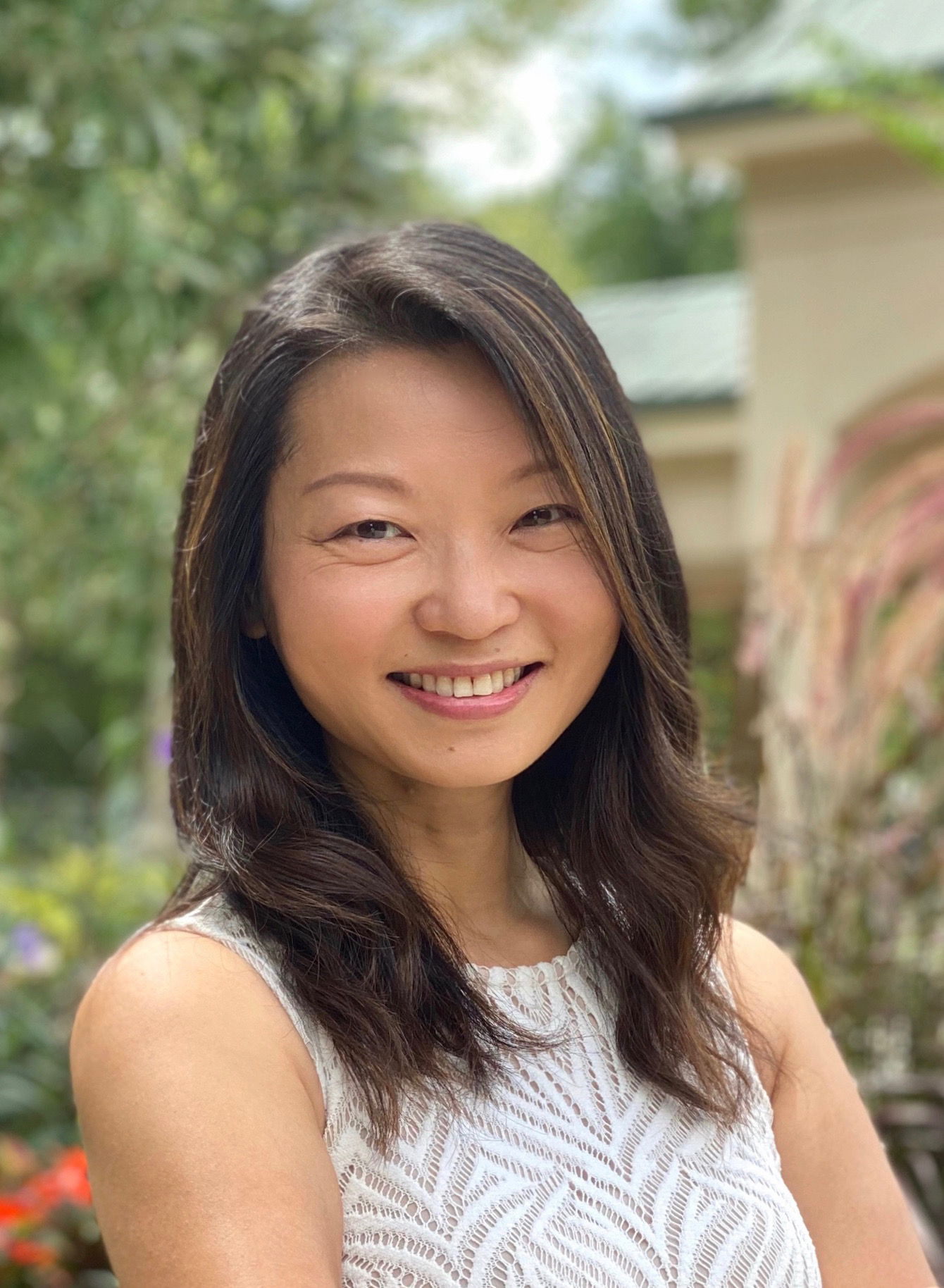}}]{Li Xiong}
Li Xiong is a Professor of Computer Science and Biomedical Informatics at Emory University. She and her research lab, Assured Information Management and Sharing (AIMS), conduct research on algorithms and methods for data management, machine learning, and data privacy and security. She has published over 160 papers and received six best paper (runner up) awards. She has served and serves as associate editor for IEEE TKDE, IEEE TDSC, and VLDBJ, general co-chair for CIKM 2022 and program co-chair for ACM SIGSPATIAL 2018, 2020, IEEE Big Data 2020. She is an IEEE fellow.
\end{IEEEbiography}

\vfill\eject
\end{document}